\documentclass[sigconf,author]{acmart}
\usepackage[linesnumbered,lined,vlined,ruled]{algorithm2e}
\usepackage{amssymb}

\usepackage{times}
\usepackage{latexsym}
\usepackage{url}
\usepackage{multirow}
\usepackage{amssymb}
\usepackage{amsmath}
\usepackage{textcomp}
\usepackage{xspace}
\usepackage{array}
\usepackage{subfigure}
\usepackage{enumitem}
\usepackage{algorithmic}

\newcommand\dc{\emph{Diagnostic Captioning}\xspace}
\newcommand\dt{\emph{Diagnostic Tagging}\xspace}

\newcommand\imageclefcaption{\textsc{ICLEFcaption}\xspace}\newcommand\peirgross{\textsc{PEIR Gross}\xspace}

\newcommand\iuxray{\textsc{IU X-Ray}\xspace}
\newcommand\mimic{\textsc{MIMIC-CXR}\xspace}

\newcommand\onenntag{\textsc{CNN+NN}\xspace}
\newcommand\onennplus{\textsc{RTEx@X}\xspace}
\newcommand\knntag{\textsc{CNN+kNN}\xspace}

\newcommand\binarycxn{\textsc{RTEx@R}\xspace}
\newcommand\taggingcxn{\textsc{RTEx@T}\xspace}
\newcommand\captioningrnn{\textsc{RTEx@X}\xspace}
\newcommand\chexpert{\textsc{CheXpert}\xspace}

\newcommand\random{\textsc{Random}\xspace}
\newcommand\snt{\textsc{S\&T}\xspace}
\newcommand\sntall{\textsc{S\&T@all}\xspace}

\newcommand\edtagprior{\textsc{EtD}\xspace}
\newcommand\edtagc{\textsc{S\&T+}\xspace}
\newcommand\rext{\textsc{RTEx}\xspace}

\newtheorem{problem}{Problem}

\setcopyright{acmcopyright}
\copyrightyear{2020}
\acmYear{2020}

\begin{document}

\title{\rext: A novel methodology for Ranking, Tagging, and Explanatory diagnostic captioning of radiography exams}

\author{Vasiliki Kougia and John Pavlopoulos* and Panagiotis Papapetrou}
\affiliation{\institution{Stockholm University, Sweden}}
\email{(kougia.vasiliki,ioannis,panagiotis)@dsv.su.se}

\author{Max Gordon}
\affiliation{\institution{Karolinska Institutet, Sweden}}
\email{max.gordon@ki.se}

\begin{abstract}
This paper introduces \rext, a novel methodology for a) ranking radiography exams based on their probability to contain an abnormality, b) generating abnormality tags for abnormal exams, and c) providing a diagnostic explanation in natural language for each abnormal exam. The task of ranking radiography exams is an important first step for practitioners who want to identify and prioritize those radiography exams that are more likely to contain abnormalities, for example, to avoid mistakes due to tiredness or to manage heavy workload (e.g., during a pandemic). We used two publicly available datasets to assess our methodology and demonstrate that for the task of ranking it outperforms its competitors in terms of $ndcg@k$. For each abnormal radiography exam \rext generates a set of abnormality tags alongside an explanatory diagnostic text to explain the tags and guide the medical expert. Our tagging component outperforms two strong competitor methods in terms of $F1$. Moreover, the diagnostic captioning component of \rext, which exploits the already extracted tags to constrain the captioning process, outperforms all competitors with respect to clinical precision and recall. 
\end{abstract}


\maketitle

\section{Introduction} \label{intro}
\begin{figure}[ht]
\centering
\includegraphics[width=0.45\textwidth]{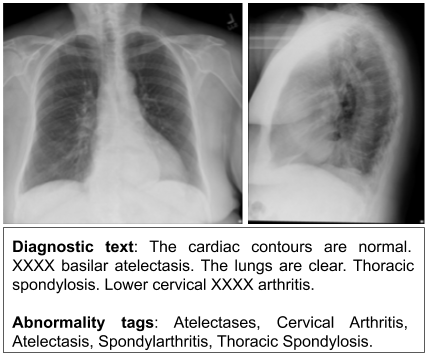}
\caption{A PA/lateral chest radiography exam along with the corresponding human-authored \textsc{diagnostic text} from IU X-Ray, and the abnormality tags. The `XXXX' is due to the de-identification process.}
\label{fig:iu_xray_case}
\end{figure}

Medical imaging is the method of forming visual representations of the anatomy or a function of the human body using a variety of imaging modalities (e.g., CR, CT, MRI) \citep{suetens2009fundamentals,aerts2014decoding}. In this paper, we particularly focus on chest radiography exams, which contain medical images produced by X-Rays. It is estimated that over three billion radiography exams are performed annually worldwide \citep{krupinski2010current}, making the daily need for processing and interpretation of the produced radiographs paramount. An example of a radiography exam is provided in Fig.~\ref{fig:iu_xray_case}, that consists of two chest radiographs together with the diagnostic text, describing the medical observations on the radiographs, and a list of abnormality tags indicating most critical observations in the exam. 
In the diagnostic text, we observe that two findings are normal (i.e., cardiac contours and lungs), while three are abnormal, i.e., \emph{thoracic spondylosis}, \emph{lower cervical arthritis}, and \emph{basilar atelectasis}. These abnormal findings are also consistent with the abnormality tags. 

Our main objective in this paper is to introduce a novel methodology for automated and explainable \dt of a collection of radiography exams, each comprising several radiographs, that can (1) accurately rank exams with abnormalities included in the radiographs, (2) automatically provide tags corresponding to the medical findings of the abnormal exams, (3) produce a diagnostic text describing the abnormality findings by exploiting both radiographs and the generated tags.  
Despite the importance of this problem, existing solutions are hindered by three major challenges.

\smallskip
\noindent \textbf{Challenge I - Screening and prioritization.}  The daily routine of diagnostic radiologists includes the examination of radiographs, i.e., medical images produced by X-Rays, for abnormalities or other findings, and an explanation of these findings in the form of a medical report per radiography exam \cite{Monshi2020}. This is a rather challenging and time-consuming task imposing a high burden both to radiologists and patients. For example, approximately 230,000 of patients in England are waiting for over a month for their imaging test results \cite{Royal2016}, while 71\% of the clinics in the U.K. report a lack of clinical radiologists \cite{Royal2019}. While several methods have emerged that automatically detect abnormalities in radiographs \cite{Pelka2019} or generate a diagnostic text \cite{Jing2018,Li2019,Liu2019}, little emphasis has been given on case prioritization and screening. 
There is hence a need for a new diagnostic approach that can \emph{automatically screen radiography exams with abnormalities} and \emph{prioritize} those with higher probability of containing an abnormality.

\smallskip
\noindent \textbf{Challenge II - Clinically correct diagnostic captioning.} 
Methods that can automatically generate (or retrieve) diagnostic text can be used to assist inexperienced physicians while they can also yield a draft to speed up the authoring process \cite{Kougia2019,Liu2019,Li2019}. However, to the best of our knowledge, all the diagnostic captioning models suggested in the literature are not optimized in terms of clinical correctness, mainly because they are trained on both normal and abnormal radiography exams. This makes them less effective compared to being trained only on abnormal exams, as we also demonstrate in Sec. \ref{ssect:cap}.  
There is hence a need for a \emph{diagnostic captioning approach} 
that is \emph{optimized for captioning abnormal radiographs}.

\smallskip
\noindent \textbf{Challenge III - Explainability\footnote{Explicitly required by EU's General Data Protection Regulation (GDRP, Art. 13§2.f: \url{gdpr.eu/article-13-personal-data-collected/}} and clinical relevance} are often provided in the form of visual highlights (e.g., heatmaps) alongside diagnostic tags \cite{Karim2020}. Nonetheless, system-generated visual explanations only function as means for highlighting image parts relevant to the diagnostic tags, without any textual explanation. 
On the other hand, \dc methods can provide both a diagnosis and an explanation for the problem at hand, since they provide a whole text instead of a tag or a label. Nonetheless, the produced reports are typically of low clinical correctness, as they are not particularly optimized in terms of clinical relevance \cite{Bluemke2020}.  
The above deficiencies could be addressed by a diagnostic tagging approach that first produces tags for abnormal radiographs, and then \emph{employs the generated tags for providing clinically relevant explanations} in the form of \emph{diagnostic text}.

\smallskip
\noindent \textbf{Contributions. }
This paper addresses the aforementioned challenges with the main contributions summarized as follows:
\setlist{nolistsep}
\begin{itemize}
\item \textbf{Novelty. } We introduce \rext, a novel methodology for explainable diagnostic tagging of radiography exams, that addresses the aforementioned challenges with the help of three key functionalities: (1) \textbf{Ranking of abnormal radiography exams: } a ranking approach is employed for prioritizing exams likelier to include an abnormality from a large collection of normal and abnormal radiography exams;
(2) \textbf{Diagnostic tagging: } a tag generator is employed for generating a set of abnormality tags for the highly ranked radiography exams, trained on an independent set of abnormal radiographs;
(3) \textbf{Diagnostic captioning: } the extracted tags are finally used by \rext to generate (or retrieve) a diagnostic text, in natural language, that provides a clinically relevant explanation of the detected abnormal findings.
\item \textbf{Applicability and efficiency. } We provide an empirical evaluation of the proposed methodology, using two publicly available datasets of radiography exams \cite{DF2015,johnson2019}. Our experimental benchmarks assess the performance of \rext on the ability to (a) rank abnormal radiography exams higher than normal ones, (b) produce the correct medical abnormality tags for abnormal radiography exams, and (c) explain the reasoning behind the selection of the detected tags in the form of diagnostic text. Moreover, a runtime experiment demonstrates the time efficiency of \rext, showing that it requires only 19.78 seconds to rank 500 radiography exams, and 19.43 seconds for tagging and diagnostic captioning of the top-100 ranked exams.
\item \textbf{Effectiveness and clinical accuracy. } Our experiments demonstrate the effectiveness of \rext against state-of-the-art competitors for the tasks of ranking and tagging. Our findings additionally suggest that diagnostic captioning using the tags produced by \rext can provide more clinically accurate diagnostic text compared to not using the generated tags.  
\end{itemize}

\smallskip
\noindent The remainder of this paper is organized as follows: in Sec. \ref{sec:related} we outline the related work,
while in Sec.~\ref{sect:methods} we describe  \rext. In Sec.~\ref{sect:results} we introduce the datasets used for our empirical evaluation, we provide the experimental setup and report our results. Finally, Sec. \ref{sec:conclusions} concludes the paper and provides directions for future work.


\section{Related work}
\label{sec:related}
In this section, we outline the main body of related work on medical image ranking, medical image tagging, and diagnostic captioning. To the best of our knowledge, while many earlier works have targeted these problems individually, there is yet no comprehensive methodology for combining these three tasks with focus on radiography exams that contain abnormalities.

Automated screening of radiography exams is not a novel idea \cite{Taguchi2010,Jaeger2013,Oliveira2008}. When the number of exams is overwhelming, as for example during a pandemic, the employment of an automated system to exclude normal cases can lead to faster treatment of abnormal cases. Recently, pre-trained deep learning models, such as DenseNet-121 \cite{Huang2017} and VGG-19 \cite{Simonyan2014}, were found to discriminate well normal cases from ones with pneumonia and COVID-19 (90\% Precision and 83\% Recall) \cite{Karim2020}. The authors noted that their approach aims to ease the work of radiologists and such an assistance scenario is suggested in this work. In our solution, we also employ DenseNet-121 CNN for multi-label classification, which is considered to be the state-of-the-art \cite{Baltruschat2019}. 

Researchers have focused on labeling radiography exams that are associated with a single abnormality finding; e.g., lymph node \cite{Roth2015} or end- diastole/systole frames in cine-MRI of the heart \cite{Kong2016}. This means that an assumption is made that the problem is a priori known (e.g., abnormality related to \textsc{lymph node}). This is not always the case, for example when the radiographs of a new patient arrive for the first time to the clinic. 
Another line of research, that of exploring multiple abnormality types, has been focusing on associating medical tags (a.k.a. concepts) to radiographs which is related to content-based image retrieval (CBIR). \citet{Liu2016} trained a custom CNN to classify radiographs in 193 classes and obtained a descriptive feature vector to be further processed and used for image retrieval. Their approach was found to be more accurate than many submissions to an earlier CLEF medical image annotation challenge, but it was also inferior than the state of the art. A similar medical image annotation challenge still exists today (\imageclefcaption) with tens of submissions each year \cite{Pelka2019}.  Participating systems were asked to tag medical images extracted from open access biomedical journal articles of PubMed Central,\footnote{\url{https://www.ncbi.nlm.nih.gov/pmc/}} where the tags were automatically extracted from each figure caption using QuickUMLS \citep{Soldaini2016}. Not very highly ranked systems used engineered visual features (Scale-Invariant Feature Transform) to encode the images (26th/49), while systems using CNNs to encode the images were better placed. The 4th best system was a ResNet-101 CNN followed by an attentional RNN multi-label image classifier. The 3rd best system was a DenseNet-121 CNN encoder followed by a K-NN image retrieval system, while the 1st place was awarded to a DenseNet-121 CNN followed by a Feed Forward Neural Network classifier. This work builds on top of the two best performing systems \footnote{The 2nd place was awarded to an ensemble of the two best performing systems.}.

\dc has not yet been investigated in the literature as an explainability step of diagnostic tagging. While \citet{Gale2018} suggested the use of captioning as an explanation step, they manually assembled sentence templates for systems to learn to fill. A dataset that comprised 
medical images and texts
was introduced for a challenge \cite{ImageCLEF2017,ImageCLEF2018}, but it was very noisy (images were figures extracted from scientific articles and the gold reports were their captions) \cite{Kougia2019}. 
\dc methods are usually Encoder-Decoders \citep{Hossain2019, Bai2018, Liu2019survey}, which often originate from \emph{Generic Image Captioning}. 
Although different variations have been suggested in the literature \cite{Zhang2017b,Li2018,Liu2019,Jing2018}, most of these methods extend the very well-known Show \& Tell (\snt) model \cite{Vinyals2015} with hierarchical decoding \cite{Jing2018}, elaborate visual attention mechanisms \cite{Wang2018}, or reinforcement learning \cite{Li2018}. \snt comprises a CNN to encode the image and uses the visual representation to initialize a decoding LSTM. We employed this model to generate diagnostic text, having extended it to also encode the tags along with the image. 
\citet{Li2018} employ an Encoder-Decoder approach to either generate or retrieve the diagnostic text from a medical image. Their hybrid approach initially uses a DenseNet \cite{Huang2017} or a VGG-19 \cite{Simonyan2014} CNN to encode the image. The encoded image is used through an attention mechanism \cite{Lu2017,Xu2015} in a stacked RNN that generates sentence embeddings, each of which is used along with the encoded image by another word-decoding RNN to generate the words of the sentence. Each sentence embedding is provided as input to a Feed Forward Neural Network (FFNN) which outputs a probability distribution over a number of fixed sentences and a word decoder. If a fixed sentence has the highest probability, then this sentence is retrieved as the next sentence instead of using the word-decoding RNN. For the explanation stage of our methodology, we also experiment with CNN-RNN Encoder-Decoder methods but mainly to explain the extracted diagnostic tags, while the Encoder-Decoders are trained only on abnormal studies, which makes sentence retrieval redundant. 

\section{The \rext methodology}
\label{sect:methods}

We present \rext, a three-stage novel methodology for ranking and explainable diagnostic tagging of radiography exams, with an overview of the whole pipeline depicted in Fig.~\ref{fig:abrank}. First, we provide the problem formulation presenting the three sub-problems, addressed by each stage of \rext.

\subsection{Formulation}
Let $\mathcal{S}=\{S_1, \ldots,  S_n\}$ be a set of $n$ radiography exams, where each exam $S_i\in \mathcal{S}$ is a set of radiographs, i.e., $S_i=\{M_{i1}, \ldots, M_{im}\}$. In our target application, we have $m=2$, that is, each $S_i$ is a pair of radiographs (one frontal and one lateral). Our formulation and approach can, however, be generalized to contain an arbitrary number of radiographs $m$. 

Assume an alphabet of abnormality tags $\mathcal{A}$.
Each radiography exam $S_i$ is assigned with a set of labels $L_{i}\in \mathcal{A}$,
either listing the abnormalities that are detected in the image
or returning an empty list indicating that the image contains no abnormalities.

Based on the above, the first objective of \rext can be formulated as follows:
\begin{problem} \textbf{(radiography exam ranking)}
Given a set of radiography exams $\mathcal{S}$, a ranking function $r(\cdot)$ and an integer $k$, identify the set $\mathcal{H}^k$ of the top $k$ \underline{abnormal} exams in $\mathcal{S}$ such that $r(\cdot)$ is maximized.
\end{problem}

Next, given the retrieved set $\mathcal{H}^k$ of the top $k$ exams, our goal is to produce a set of abnormality tags. This brings us to the second objective of \rext, which can be formulated as follows:
\begin{problem} \textbf{(abnormal radiography exam diagnostic tagging)}
Given a set of abnormal radiography exams $\mathcal{H}^k$, produce a set of abnormality tags $\mathcal{T}$, with each tag originating from set $\mathcal{A}$. Each set of tags $T_j \in \mathcal{T}$ describes each exam $S_j\in \mathcal{H}^k$.
\end{problem}
In other words, all images contained in a single radiography exam ($S_j$) are described by a common set of abnormality tags ($T_j$).

Eventually, given the set of produced tags, our final goal is to obtain a diagnostic caption explaining the abnormalities shown in the images contained in the radiography exam, and referenced by the extracted tags. More formally, \rext's third objective is formulated as follows:

\begin{problem} \textbf{(abnormal radiography exam diagnostic captioning)}
Given a set of abnormal radiography exams $\mathcal{H}^k$ and a set of tags $\mathcal{T}$ describing the abnormalities in each exam, provide a set of captions $\mathcal{C}$, where each caption $C_j\in \mathcal{C}$ describes radiography exam $S_j \in \mathcal{H}^k$.
\end{problem}

\subsection{The three stages of \rext}
The three stages of \rext are outlined in Alg. 1. Next, we provide more details for each stage.
\begin{figure*}
    \centering
    \includegraphics[width=\textwidth]{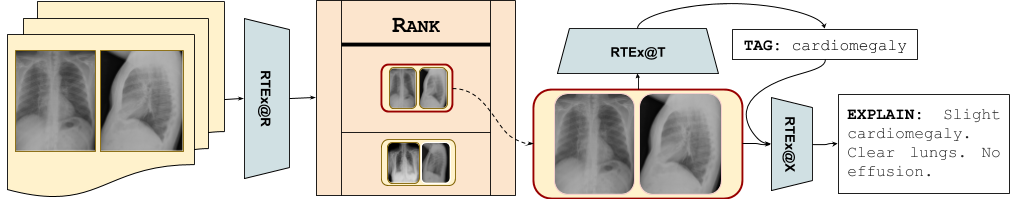}
    \caption{A depiction of our \rext methodology. First, it ranks the radiography exams based on their probability (i.e., using the radiographs of each exam) to include an abnormality. The highest ranked are tagged with abnormality terms and an explanatory diagnostic text is automatically provided to assist the expert.} 
    \label{fig:abrank}
\end{figure*}

\begin{algorithm}
\label{methodology}
\caption{Outline of the \rext methodology}
\SetAlgoVlined
    \KwData{a set of radiography exams $\mathcal{S}$ and the number $k$ of exams to retrieve.}
    \KwResult{a set $\mathcal{T}$ of abnormality tags and a set $\mathcal{C}$ of captions.}
    \text{// define a list to maintain the score of each radiography exam}\;
    $scores=\{\}$ \;
    \text{// apply the \rext ranking function}\;
    \For{$S_i \in \mathcal{S}$} {
        $scores_{i} = \binarycxn(S_i)$ \;
    }
    \text{// sort S with respect to their scores in descending order}\;
    $\mathcal{S}'= sort(\mathcal{S}, scores, ``descend")$ \;
    \text{// filter the top k abnormal exams}\;
    $\mathcal{H}^k= filter(\mathcal{S}',k)$ \;
    $\mathcal{C}, \mathcal{T} \leftarrow \{\}$ \;
    \For{$S_j \in \mathcal{H}^k$}{
        \text{// apply the \rext tagging function}\;
        $T_{j} = \taggingcxn(S_j)$ \;
        \text{// apply the \rext captioning function}\;
        $C_{j} = \captioningrnn(S_j, T_j)$ \;
    }
    \KwRet $\{\mathcal{T}, \mathcal{C}\}$
\end{algorithm}

\subsubsection{\binarycxn: Ranking}
\label{methods:abnorank}
\begin{figure}[h]
    \centering
    \includegraphics[width=0.5\textwidth]{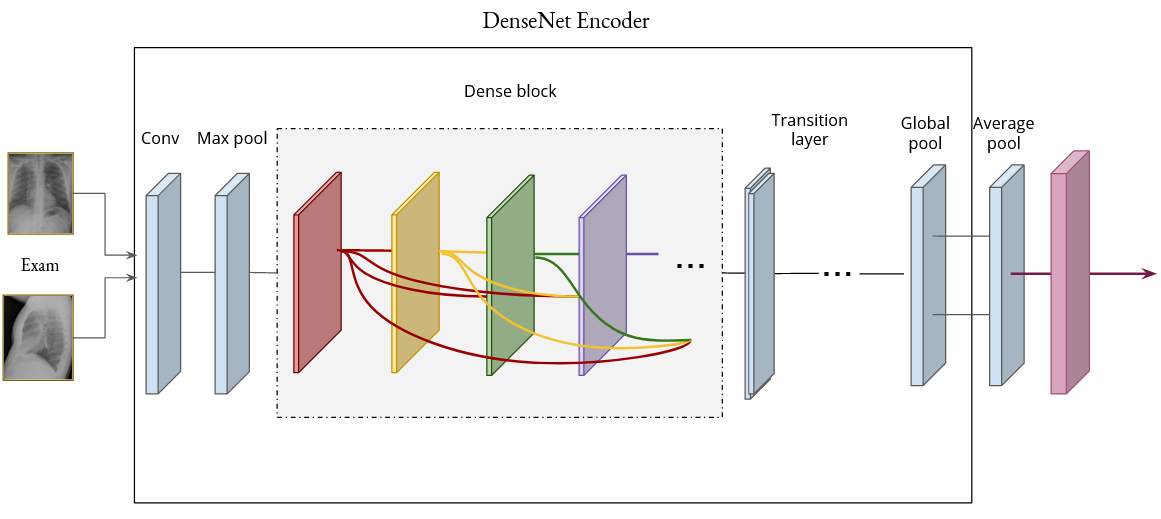}
    \caption{The architecture of \binarycxn. The input is a radiography exam and the output is a probability of the exam to be abnormal.}
    \label{fig:bicxn}
\end{figure}

For the first stage in our methodology we implement an architecture which we refer to as \binarycxn, shown in Fig.~\ref{fig:bicxn}. More concretely, we employ the same visual encoder as in \cite{Rajpurkar2017}. That is the DenseNet-121 CNN, which is followed by a Feed Forward Neural Network (FFNN). The input of the network are images of radiography exams while the output is a score representing the probability that the exam in question is \emph{abnormal}. First, both images of the exam are fed to DenseNet-121 (depicted inside the box in the center) and an embedding for each image is extracted from its last average pooling layer. These embeddings are concatenated to yield a single embedding for the radiography exam. Then, the exam embedding is passed to a FFNN with a $sigmoid$ to return a score from 0 (normal) to 1 (abnormal).

\subsubsection{\taggingcxn: Diagnostic tagging} 
\label{methods:abnotag}
\begin{figure}[h]
    \centering
    \includegraphics[width=0.5\textwidth]{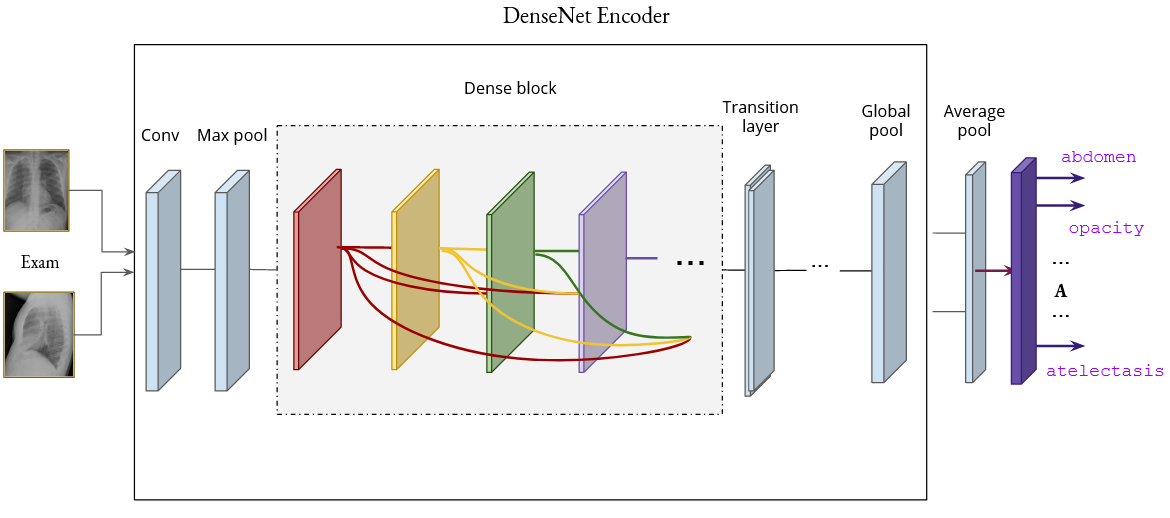}
    \caption{The architecture of \taggingcxn, which is similar to \binarycxn, but the input is an abnormal radiography exam and the output consists of $A$ binary nodes, where $A$ is the total number of tags in the dataset. The nodes that yield probabilities higher than a defined threshold, indicate the presence of the respective medical abnormalities.}
    \label{fig:tagcxn}
\end{figure}

The second stage of our methodology, referred to as \taggingcxn  comprises the assignment of a set of tags $T_j$ to a radiography exam $S_j \in H^h$. Our method for addressing this task is called \taggingcxn and shown in Fig.~\ref{fig:tagcxn}. It is similar to \binarycxn in that it uses the DenseNet-121 CNN encoder and a FFNN. But it differs in that the FFNN has one output and one sigmoid activation per abnormality tag in the dataset, leading to $A$ different output nodes (the right most arrows in the figure). In effect, it returns a probability distribution over the abnormality tags and if the probability of an abnormality tag (i.e., its respective node) exceeds a learned threshold, then the tag is assigned to the radiography exam. 

\subsubsection{\captioningrnn: Diagnostic captioning} \label{methods:captioning}
For the last stage of our methodology (the rightmost part of Fig.~\ref{fig:abrank}), referred to as \captioningrnn, we use a method that comprises a DenseNet-121 CNN encoder, calibrated for the task of diagnostic captioning. 
More specifically, each radiography exam in the database is encoded (offline) by our CNN to an embedding (i.e., two image embeddings extracted from the last average pooling layer of the encoder, concatenated). Our CNN also encodes any new test exam. Then, the cosine similarity between the test embedding and all the training embeddings in the database are calculated and the most similar exam is retrieved from the database. Its diagnostic text is then assigned to the test exam. \onennplus limits its search to training exams that have the exact same tags as the ones predicted (during the tagging stage) for the test exam. However, the whole database is searched, when no exams exist with the same tags.
We note that all the embeddings are first normalized (using $L2$), so that the cosine similarity between a test embedding and all the training embeddings in the database is computed with a single matrix (element wise) multiplication. This reduces the search time from minutes to milliseconds, making this method in effect the most efficient compared to its competitors.

\section{Empirical evaluation}
\label{sect:results}
In this section, we describe the datasets used for our experiments, we provide details on the experimental setup, and present our results.

\subsection{Datasets} \label{sect:data}
Datasets that can be used for \dc comprise medical images and associated diagnostic reports \footnote{We limit our radiography exams to datasets with reports in English.}. We are aware of four such publicly available datasets. Namely, \iuxray, \mimic, \peirgross and \imageclefcaption, but we employ only the former two, which are of high quality \citep{Kougia2019}.\footnote{\peirgross comprises medical images and photographs, mainly for educational purposes. \imageclefcaption comprises images extracted from scientific articles and uses the caption of each such image as the respective report.} 

\smallskip
\noindent\textbf{\iuxray:}
\label{ssec:iuxray}
IU X-ray \citep{DF2015} is a collection of radiology exams, including chest radiographs, abnormality tags, and radiologist narrative reports, and is publicly available through the Open Access Biomedical Image Search Engine (OpenI).\footnote{\url{https://openi.nlm.nih.gov/}} The dataset consists of 3,995 reports (one report per patient) and 7,470 frontal or lateral radiographs, with each radiology report consisting of an `Indication' (e.g., symptoms), a `Comparison' (e.g., previous information about the patient), a `Findings' and an `Impression' section. Each report contains two groups of tags. First, there are manual tags\footnote{A combination of MeSH (\url{https://goo.gl/iDvwj2}) and Radiology Lexicon codes (\url{http://www.radlex.org/}).} assigned by two trained coders, each comprising a heading (disorder, anatomy, object, or sign) and subheadings (e.g., `Hiatal/large', where `large' indicates the anatomical site of the disease). Second, the `Findings' and `Impression' sections were used to associate each report with a number of automatically extracted tags, produced by Medical Text Indexer \citep{Mork2013} (MTI tags). An example case is that shown in Fig.~\ref{fig:iu_xray_case}, where it can be seen that the MTI tags are simple words or terms (e.g., `Hiatus'). 

For the ranking stage of our methodology, each exam was labeled as \emph{abnormal}, if one or more manual abnormality tags were assigned, and \emph{normal}, otherwise (the tag `normal' or `no indexing' was assigned). For the tagging stage of our methodology, we employed the MTI codes, because the manual codes do not explicitly describe the abnormality, but most often also include other information (e.g., anatomical site). For the explanation stage, we employed the `Findings' section. Also, in our experiments we used only exams with two images considering this to be the standard (one frontal and one lateral radiograph), and excluded the rest. We also discarded the exams that did not have a `Findings' section. This resulted in 2,790 exams, from which 1,952 are used for training, 276 for validation and 562 for testing.\footnote{We used the same split as in \citet{Li2018,Li2019}}
The class ratio in the dataset is slightly imbalanced with 39\% normal radiology exams. Abnormal exams are assigned with 3 tags on average, while the most frequent tag is `degenerative change' (216 exams).
The length of the diagnostic text in each report is 40 words on average. For the normal exams the diagnostic text can be exactly the same for many different patients, e.g., the following finding `The heart is normal in size. The mediastinum is unremarkable. The lungs are clear.' appears in 29 exams. By contrast, the most frequent abnormal report appeared exactly the same in 7 reports.

\smallskip
\noindent\textbf{\mimic:}
\label{ssec:mimic}
This dataset comprises 377,110 chest radiographs associated with 227,835 radiography exams which come from 64,588 patients of the Beth Israel Deaconess Medical Center between 2011-2016.\footnote{MIMIC-CXR v2.0.0, \url{https://mimic-cxr.mit.edu/}} As in \iuxray, reports in MIMIC are organized in sections, but some reports include additional sections such as `History', `Examination', or `Technique', but not in a consistent manner, because the structure of the reports and the section names were not enforced by the hospital's user interface \citep{johnson2019}. The current version of the dataset does not contain the initial labels, so we re-produced them by applying the \chexpert disease mention labeler \cite{Irvin2019} on the reports as described in \citet{johnson2019}. \chexpert classifies texts into 14 labels (13 diagnoses and `No Finding'), each as 
`negative', `positive', or `uncertain' for a specific text. We treated those labeled uncertain as positive. For the ranking step, we labeled exams as normal when the `No Finding' label was assigned. In total, there are 40,306 exams with two images that correspond to 29,482 patients. After removing 11 exams that did not have a `Findings' section, which we used for the explanation stage of our \rext, we split the dataset to 70\% (training), 10\% (validation), and 20\% (test) with respect to patients. For our experiments we randomly kept one exam per patient and sampled 2,300 patients from the training set, 300 from the validation set and 650 from the test set; with 68\% of this final dataset consisting of normal exams. Each abnormal exam has 2 labels on average, while the most common label is `Pneumonia'. The average diagnostic text length is 55 words. In this dataset many normal cases have the same diagnostic text, e.g, the most common normal caption appears in 53 exams. Considering only the abnormal exams the most frequent caption appears 4 times. 

\subsection{Experimental setup}
For each of the three stages of \rext we benchmark each technique against competitors. Next, we outline the competitor methods and the performance metrics used for each benchmark.

\subsubsection{Ranking and Tagging}
We investigated one baseline method, referred to as \random, and two competitor methods, referred to as \onenntag and \knntag, for both ranking and tagging stages. The methods were benchmarked against \binarycxn and \taggingcxn, respectively. For the two competitors, the ranking is determined based on the produced tags by these methods trained on both normal and abnormal exams. Moreover, at the tagging stage, the tags are obtained by retraining the same methods only on abnormal exams. Next, we describe the baseline as well as the two tagging methods.

\smallskip
\noindent \textbf{\random. } This is a baseline method used both for ranking and tagging and simulates the case where no screening is performed. For the ranking task it randomly returns a number serving as the abnormality probability. For tagging, it simply assigns a set of random tags from the training set. The number of tags assigned is the average number of tags per training exam. 

\smallskip
\noindent \textbf{\onenntag. } This method employs a DenseNet-121 CNN \cite{Huang2017} encoder. It is pre-trained on ImageNet and fine-tuned on our datasets (\iuxray or \mimic). \onenntag encodes all images (from the training and test sets) and concatenates the obtained representations for each radiograph in an exam ($S_j$), to yield a single representation per exam ($V_j$). Then, for each test representation, the \emph{cosine} similarity against all the training representations is computed and the nearest exam is returned. When generating tags 
then the abnormality tags 
of the nearest exam are returned and assigned to the test exam. 

\smallskip
\noindent \textbf{\knntag. } This method is an extension of \onenntag that uses the $k$-most similar training exams to compute the tags $T_j$ for exam $S_j$. To constrain the number of returned tags ($|T_j|$), only the $r$ most frequent tags of the $k$ exams are held. Moreover, we set $r$ to be the average number of tags per exam of the particular $k$ retrieved exams. We observe that \knntag is considered a very strong baseline for tagging. It was ranked third in a recent medical tagging competition \cite{Kougia2019b}. The first two methods are \taggingcxn (see Section~\ref{methods:abnotag}) and an ensemble of \knntag and \taggingcxn, respectively. 

\smallskip
\noindent For solving the problem of ranking, we adapted \onenntag and \knntag as follows. The abnormality tags of the most similar radiography exam in the training set are returned and a probability score $p$ is computed using the following formula:
\begin{equation} 
\label{eq:rank}
P = \frac{\sum_{t \in T_j} rel(t)} {|\mathcal{G}|} \ ,
\end{equation}
\noindent where $\mathcal{G}$ are all the ground truth tags of the dataset, $T_j$ are the generated tags for radiography exam $S_j$ and $rel(t) = 1$ when $t \in \mathcal{G}$ and zero otherwise. $P$ will usually be close to zero. The main intuition is that the more the assigned tags, the higher the $P$ and the likelier it is that this exam is abnormal.

\smallskip
\noindent \textbf{Evaluation metrics. } 
Ranking methods were evaluated in terms of $nDCG@k$, with a varying $k$. We also used $Precision@k$, but preliminary experiments showed that this measure correlates highly with $nDCG@k$.
Tagging methods were evaluated in terms of $F1@k$. We used the top-$k$ abnormal cases (ranked by \binarycxn) to compute the F1 score between their predicted and their gold tags.

\subsubsection{Diagnostic captioning}

We benchmarked three competitors for the task of diagnostic captioning showing the benefits in terms of clinical correctness when using the generated tags.

\smallskip
\noindent \textbf{\snt} This method was introduced by \citet{Vinyals2015} for image captioning and it is only applicable for the stage of diagnostic captioning. As the encoder of the \snt architecture we employ the DenseNet-121 \cite{Huang2017} CNN, which is used to initialize an LSTM-RNN decoder \cite{Hochreiter1997}. A dense layer on top outputs a probability distribution over the words of the vocabulary, so that the decoder generates a word at a time. The word generation process continues until a special `end' token is produced or the maximum caption length is reached. 

\smallskip
\noindent \textbf{\edtagc} This method extends \snt (also applicable solely to diagnostic captioning), so that the generated text explains the predicted tags. Hence, after the encoding phase and prior to the decoding phase (before the generation of the first word), the tags are provided to the decoder, as if they were words of the diagnostic text; similar to \emph{teacher forcing} \cite{Goodfellow2016}. Since the decoder is an RNN, this acts as a prior during the decoding that will follow.

\smallskip
\noindent \textbf{\edtagprior} This method follows a tag and image constrained Encoder-Decoder architecture. A DenseNet-121 CNN \cite{Huang2017} yields one visual embedding per exam. The decoder is an LSTM constrained from the visual embedding and the tags $T_j \in T$ that were assigned to exam $H_j \in H^k$ during the previous step (see Section~\ref{methods:abnotag}). We call this method \edtagprior. More formally, the decoder at each time step $s$ learns a hidden state $h_{s}$ as the non-linear combination (the weight matrix $W$ is learned) of the input word $x_{s}$ and the previous hidden state $h_{s-1}$:
\begin{equation}
\begin{split}
\nonumber
    i_{s} &= \sigma(W_{i} \cdot [x_s, V_j, E_j, h_{s-1}] + b_{i})\\
    f_{s} &= \sigma(W_{f} \cdot [x_{s}, V_j, E_j, h_{s-1}] + b_{f})\\
    o_{s} &= \sigma(W_{o} \cdot [x_{s}, V_j, E_j, h_{s-1}] + b_{o})\\
    q_{s} &= \tanh(W_{q} \cdot [x_{s}, V_j,  E_j, h_{s-1}] + b_{q})\\
    c_{s} &= f_{s} \cdot c_{s-1} + i_{s} \cdot q_{s}\\
    h_{s} &= o_{s} \cdot \tanh(c_{s}),
\end{split}
\end{equation}
\noindent
where $i_{s}$, $f_{s}$ are the LSTM input and forget gates regulating the information from this and the previous cell to be forgotten. $V_j$ is the visual representation from the last average pooling layer of the DenseNet encoder. $E_j$ is the centroid of the word embeddings of the tags $T_j$:
$$
E_j = \frac{1}{|T_j|}\sum_{t \in T_j} W_e \cdot t
$$

\smallskip
\noindent
For all the text generation methods mentioned above, we preprocessed the text by tokenizing, lower-casing the words, removing digits and words with length 1. We used the Adam optimizer \cite{Kingma2014} everywhere with initial learning rate 10e-3. \taggingcxn and \binarycxn used a learning rate reduced mechanism \cite{Rajpurkar2017}.

\smallskip
\noindent \textbf{Evaluation metrics. }  We employed both word-overlap and clinical correctness measures to evaluate the system-produced diagnostic text. The most common word-overlap measures in diagnostic captioning are BLEU \citep{Papineni2002} and ROUGE-L \citep{Lin2004}. BLEU is precision-based and measures word n-gram overlap between the produced and the ground truth texts. ROUGE-L measures the ratio of the length of the longest common n-gram shared by the produced text and the ground truth texts, to either the length of the ground truth text (ROUGE-L Recall) or the length of the generated text (ROUGE-L Precision). We employ the harmonic mean of the two (ROUGE-L F-measure). For the implementations of BLEU and ROUGE-L, we used respectively sacrebleu\footnote{\url{https://github.com/mjpost/sacrebleu/blob/master/sacrebleu/sacrebleu.py}} and MSCOCO\footnote{\url{https://github.com/salaniz/pycocoevalcap/tree/master/rouge}}. To evaluate the clinical correctness,
following the work of \cite{Liu2019}, we used the CheXPert labeler \cite{Irvin2019} to extract labels from both the ground truth and the system-generated diagnostic texts. Clinical precision (CP) is then the average number of labels shared between the ground truth and system-generated texts, to the number of labels of the latter. Similarly, clinical recall (CR) is the average number of labels shared between the ground truth and system-generated texts, to the number of labels of the former.

\subsection{Experimental results}
\label{ssect:results}
Next, we present our results with regard to ranking, tagging, and diagnostic captioning. Finally, we provide a discussion of our findings and assess the overall performance of \rext.

\subsubsection{Ranking} 
\begin{figure}[t]
\centering
    \subfigure[MIMIC-CXR.]{%
    \includegraphics[width=.225\textwidth]{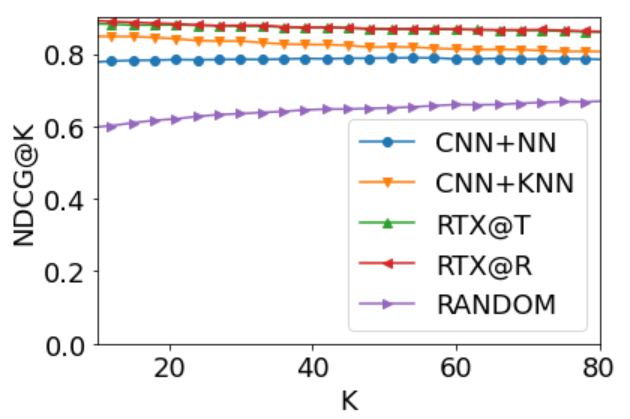}
    }
    \subfigure[IU X-ray.]{%
    \includegraphics[width=.225\textwidth]{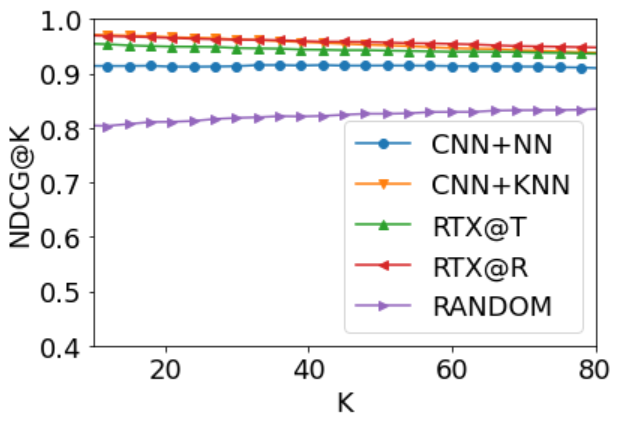}
    }
\caption{NDCG@K of all methods for the task of ranking radiography exams based on the probability of abnormality for \mimic (a) and \iuxray (b). We used bootstrapping (1000 samples of 100 exams each) and report the average value. $K$ varies from 10 to 80 and moving average was used with a window of 5. For \mimic we observe that \binarycxn and \taggingcxn consistently outperform the other methods, while for \iuxray the winners are \binarycxn and \knntag.}
\label{fig:ndcg_scores}
\end{figure}

Fig.~\ref{fig:ndcg_scores} (a) and (b) depict the performance of the methods in terms of $NDCG@K$.\footnote{Similar results were obtained in terms of $Precision@K$.} We used bootstrapping, sampling 100 exams at a time, varying $K$ from 10 to 80 radiography exams. \random is outperformed by all competitors, while \binarycxn is the overall winner for both datasets, with the second best being \taggingcxn for \mimic and \knntag for \iuxray.

\subsubsection{Diagnostic tagging}
\label{ssect:xtag}

\begin{figure}[t]
\centering
    \subfigure[MIMIC-CXR.]{%
    \includegraphics[width=.225\textwidth]{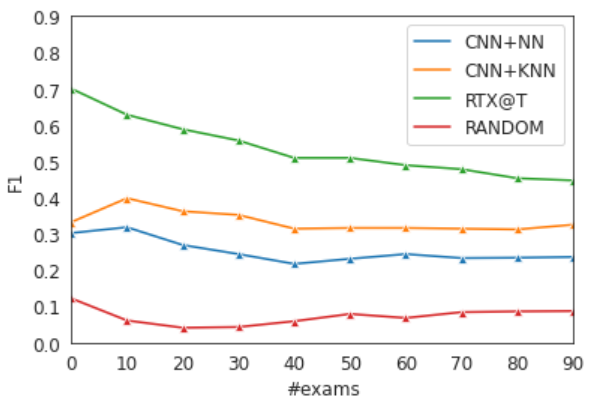}
    }
    \subfigure[IU X-ray.]{%
    \includegraphics[width=.225\textwidth]{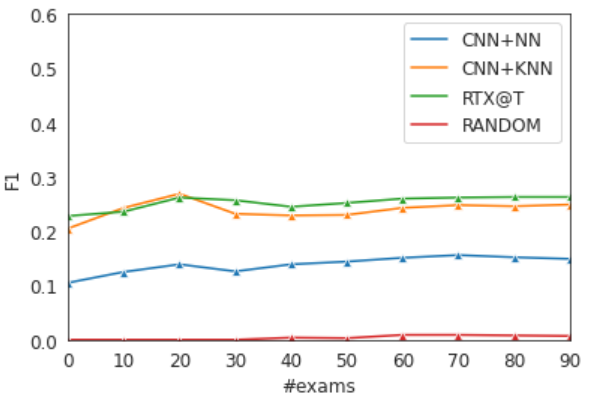}
    }
\caption{F1 of diagnostic tagging methods, on the top 100 ranked radiography exams. The cases were ranked by \binarycxn, based on their abnormality probability for \mimic (a) and \iuxray (b). We observe that \binarycxn is the winner for both datasets, with \knntag being the second best by up to a factor of two for \mimic.}
\label{fig:f1_tagging_scores}
\end{figure}

During this step we assume that the radiography exams are already ranked based on an abnormality probability. Thus, we evaluate various methods with respect to their ability to correctly detect the correct abnormality tags. We report \emph{Macro F1} (macro averaging across exams), which is also the standard measure of a recent competition on medical term tagging \cite{Pelka2019}. 
As it can be seen in Fig,~\ref{fig:f1_tagging_scores}, \taggingcxn outperforms the two competitors in both datasets, with the second best being \knntag with a difference of up to a factor of two for \mimic.

\subsubsection{Diagnostic captioning}
\label{ssect:cap}

\begin{table}[ht]
\begin{center}
\begin{tabular}{|l|l|c|c||c|c|}\hline 
\bf Dataset & \bf Model & \bf BLEU & \bf ROU & \bf CP & \bf CR \\\hline
\multirow{5}{*}{MIMIC-CXR} 
& \sntall & 7.8 & 25.7 & 0.080 & 0.118 \\ \cline{2-6}
& \snt & 8.2 & 25.2 & 0.208 & 0.151 \\ \cline{2-6}
& \edtagc & \textbf{9.8} & \textbf{26.2} & 0.081 & 0.117 \\ \cline{2-6}
& \edtagprior & 6.9 & 25.5 & 0.171 & 0.144 \\ \cline{2-6}
& \onennplus & 5.9 & 20.5 & \textbf{0.229} & \textbf{0.284} \\ \cline{2-6} 
\hline \hline
\multirow{5}{*}{IU X-ray} 
& \sntall & 6.9 & 23.6 & 0.118 & 0.088 \\ \cline{2-6}
& \snt & 6.5 & 23.0 & 0.153 & 0.113 \\ \cline{2-6}
& \edtagc & 9.5 & 23.4 & 0.085 & 0.071 \\ \cline{2-6}
& \edtagprior & \textbf{10.0} & \textbf{26.7} & 0.131 & 0.124 \\ \cline{2-6}
& \onennplus & 5.5 & 20.2 & \textbf{0.193} & \textbf{0.222} \\ \cline{2-6}
\hline
\end{tabular}
\end{center}
\caption{The results of our explanatory captioning phase, evaluated with BLEU, ROUGE-L (ROU), Clinical Precision (CP) and Clinical Recall (CR). Clinical correctness decreases when \snt is trained also on normal exams (\sntall). Our \onennplus outperforms all other methods in clinical precision and recall.}
\label{tab:captioning}
\end{table}

Table~\ref{tab:captioning} provides the results of the methods for the task of \dc. We considered as \emph{ground truth}, i.e., set $\mathcal{G}$, the correct reports and as \emph{predicted captions} the system-produced diagnostic texts. Our \onennplus outperforms all methods in terms of clinical precision and recall. Generative models achieve higher word-overlap scores, mainly because they learn to repeat common phrases that exist in the reports. On the other hand, retrieval methods assign texts that are written from radiologists, so they have a higher clinical value.
When training \snt on all exams (\sntall), using both normal and abnormal cases, clinical precision and recall decrease in both datasets. By contrast, the performance in terms of word-overlap measures (BLEU and ROUGE-L) was slightly improved overall, probably because the decoder is now better in generating text present in normal reports, which however is also present in abnormal reports (see Fig.~\ref{fig:iu_xray_case}).

\subsubsection{Runtime}
\label{ssect:runtime}
As a final benchmark we calculated the runtime of \rext on ranking, tagging, and captioning on 500 randomly selected radiography exams from our IUXray test set. Ranking lasted 19.78 seconds. Producing tags and diagnostic texts for the top 100 ranked exams lasted 19.43 seconds. Nonetheless, all 100 top-ranked exams in this experiment were abnormal. Note that an experienced radiologist needs 2 minutes on average \cite{Royal2019} for reporting a radiography exam, hence 200 minutes for 100 exams. The experiment was performed on a 32-core server with 256GB RAM and 4GPUs.

\smallskip
\noindent
\textbf{Repeatability.} For repeatability purposes, the code for the best performing pipeline of \rext is available on github.\footnote{\url{https://github.com/ipavlopoulos/rtex.git}}

\section{Conclusions}
\label{sec:conclusions}
We introduced a new methodology that can be used for (1) ranking radiography exams based on the probability of containing an abnormality, (2) producing diagnostic tags using abnormal exams for training, and (3) providing diagnostic text produced based on both the radiographs and tags, as means of explaining the predicted tags. This is an important step for practitioners to prioritize cases with abnormalities. Our methodology can be further used to predict abnormality tags and complement them with an automatically suggested explanatory diagnostic text to guide the medical expert. We experimented with two publicly available datasets showing that our ranking and tagging 
components outperform two strong competitors and a baseline. Our diagnostic captioning component demonstrates the benefit of employing tags for generating text of higher clinical correctness. We also demonstrated that limiting our training data to only abnormal exams improves the clinical correctness of the automatically provided text.
Future directions include further experimentation with data of a larger scale and deployment to hospitals.

\bibliographystyle{ACM-Reference-Format}
\bibliography{paper}

\end{document}